\documentclass{article}

\usepackage{arxiv}

\usepackage[utf8]{inputenc} 
\usepackage[T1]{fontenc}    
\usepackage{hyperref}       
\usepackage{url}            
\usepackage{booktabs}       
\usepackage{amsfonts}       
\usepackage{nicefrac}       
\usepackage{microtype}      
\usepackage{lipsum}
\usepackage{subcaption}
\usepackage{graphicx}
\usepackage{placeins}
\usepackage{epsfig}
\usepackage{amsmath}
\usepackage{color}
\usepackage{wrapfig}

\title{Stochasticity in Neural ODEs: An Empirical Study}

\author{
  Viktor Oganesyan\textsuperscript{*} \thanks{Samsung-HSE Laboratory, National Research University Higher School of Economics} \\
  \And
 Alexandra  Volokhova\textsuperscript{*}\footnotemark[1]~ \thanks{Moscow Institute of Physics and Technology} \\
  \And
 Dmitry Vetrov\footnotemark[1\;]~ \thanks{Samsung AI Center Moscow}\\
}

\begin{document}
\maketitle

\begin{abstract}

Stochastic regularization of neural networks (e.g. dropout) is a wide-spread technique in deep learning that allows for better generalization.
Despite its success, continuous-time models, such as neural ordinary differential equation (ODE), 
usually rely on a completely deterministic feed-forward operation.
This work provides an empirical study of stochastically regularized neural ODE on several image-classification tasks (CIFAR-10, CIFAR-100, TinyImageNet).
Building upon the formalism of stochastic differential equations (SDEs), we demonstrate that neural SDE is able to outperform its deterministic counterpart.
Further, we show that data augmentation during the training improves the performance of both deterministic and stochastic versions of the same model.
However, the improvements obtained by the data augmentation completely eliminate the empirical gains of the stochastic regularization, making the difference in the performance of neural ODE and neural SDE negligible.

\end{abstract}

\section{Introduction}

Deep neural networks describe an expressive parametric family of functions, which can be used for building predictive models in many machine learning tasks.
However, the expressive power of neural networks comes at the cost of potential overfitting on the training data.
To prevent this undesired behavior many regularization techniques have been developed and successfully applied (l2-regularization, dropout, early stopping, batch normalization) \cite{murphy2012machine, srivastava2014dropout, ioffe2015batch}.

Recently, continuous-time neural models (neural ODE \cite{chen2018neural}) have attracted the attention of the community.
In contrast to the conventional deep learning models, they operate by parameterizing an ordinary differential equation (ODE) with a neural network.
This approach is demonstrated to be promising in the design of normalizing flows \cite{grathwohl2018ffjord} and time-series generative models \cite{rubanova2019latentode, yildiz2019ode2vae, jia2019neuraljumpsde, li2020scalablesde}
Moreover, continuous-time models directly generalize ResNet architecture \cite{he2016deep}, which is known to be efficient in practice by preventing gradient saturation and allowing for performance gains with an increase of the depth.
For supervised learning (e.g. image classification), this generalization promises such benefits as parameter-efficiency and adaptive computational time \cite{chen2018neural}.
Despite the numerous theoretical benefits, continuous-time models lack empirical analysis and design guidelines, which significantly hinders the development of novel models and their application to practical tasks.

\begin{figure}
\begin{subfigure}{.25\textwidth}
  \centering
  \includegraphics[width=0.94\linewidth]{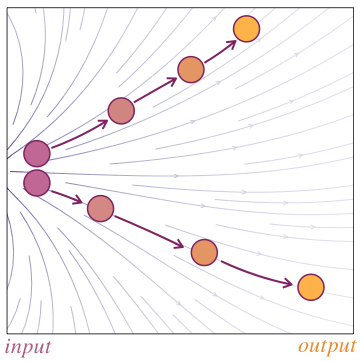}
  \caption{}
  \label{fig_1:divergent_field}
\end{subfigure}%
\begin{subfigure}{.25\textwidth}
  \centering
  \includegraphics[width=\linewidth]{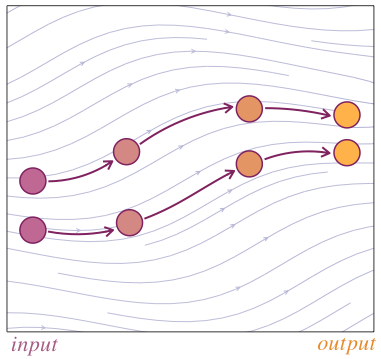}
  \caption{}
  \label{fig_1:gentle_field}
\end{subfigure}
\begin{subfigure}{.5\textwidth}
  \centering
  \includegraphics[width=0.956\linewidth]{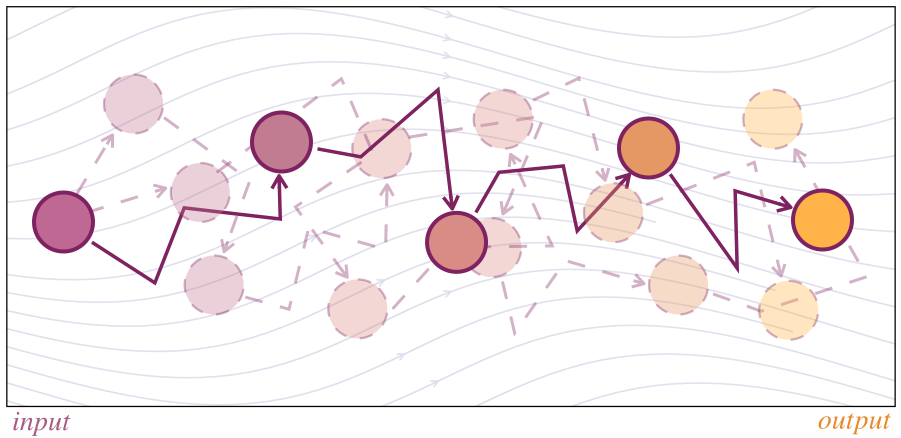}
  \caption{}
  \label{fig_1:stochastic}
\end{subfigure}
\caption{
Illustrative trajectories of integration inside the neural ODE (a, b) and the neural SDE (c). Neural ODE performs an integration on a forward pass according to a dynamic function defined by a neural net. Thus, the input of this model follows some trajectory during the integration. Figure (a) shows trajectories into the neural ODE with the dynamic function, which causes a highly divergent vector field. In this case, similar inputs are mapped into significantly different outputs. In contrast to mapping performed by the neural ODE with non-divergent vector field (b), where integration preserves similarity of inputs. Plot (c) illustrates the stochastic nature of trajectories inside the neural SDE. There, the input passes a random trajectory from a set of possible ones, while the model learns to make a correct prediction. We assume this stochasticity  encourages the vector field of neural SDE to be less divergent, which leads to better robustness and prevents overfitting.}
\label{fig_1}
\end{figure}

In this paper, we provide an empirical study of the stochastic regularization of neural ODE. An essential way to introduce stochastic into a neural ODE is to extend it to a neural stochastic differential equation (neural SDE) \cite{li2020scalablesde, jia2019neuraljumpsde, tzen2019theoreticalsde, liu2019neuralsde, tzen2019neuralsde}.
The intuitive motivation for this extension is the following.
As well as in conventional deep learning models, putting noise on the intermediate representations helps us to foster the generalization abilities of the model.
An overfitted continuous-time model could be represented as a highly divergent vector field, where small perturbations of initial conditions may result in completely different end-points of the dynamics (see Fig. \ref{fig_1:divergent_field}).
In contrast, the continuous-time model that generalizes well should map close initial points to similar outputs guaranteeing the robustness to small perturbations (see Fig. \ref{fig_1:gentle_field}). 
The input of neural SDE follows one random trajectory from a set of neighboring ones (Fig.\ref{fig_1:stochastic}). Whilst the model learns to predict a correct answer for the input regardless of which particular trajectory it follows. 
This encourages neural SDE to learn that neighboring trajectories should lead to close outputs, which actually prevents divergence. Thus, introducing the stochasticity at the training stage, we foster the model to learn such parameters that allow for the robust feed-forward procedure.
\let\thefootnote\relax\footnote{\textsuperscript{*}Equal contribution. Correspondence to: Alexandra Volokhova \texttt{avolohova@hse.ru}, Viktor Oganesyan \texttt{voganesyan@hse.ru}\\Workshop on Integration of Deep Neural Models and Differential Equations, ICLR 2020}
\newpage

We study continuous models on three image classification tasks: CIFAR-10, CIFAR-100 \cite{krizhevsky2009cifar}, TinyImageNet \cite{tinyimagenet}.
As a starting point of our study, we compare neural ODE with ResNet. We put both models in equal conditions (in terms of architecture and regularization) and observe that they perform similarly. 
Further, we introduce stochasticity into the neural ODE using the formalism of stochastic differential equations. 
We find out that this procedure can regularize neural ODE. 
However, our experiments show that common data augmentation allows neural ODE to achieve better generalization than the introduction of stochasticity. 
Therefore, we see that perturbing representations with data augmentation is enough for learning a robust model that generalizes well.

\section{Considered models}\label{sec:models}

We conduct experiments with three types of models: residual network, neural ordinary differential equation, neural stochastic differential equation. All these models can be regarded as the integration of a differential equation with some integration scheme.
One block of a residual network performs the following mapping:
\begin{equation}\label{eq:resnet}
    z_{out} = z_{in} + f_\theta(z_{in}),
\end{equation}
there $f_\theta$ is a neural network with parameters $\theta$, $z_{in}$ and $z_{out}$ are an input and an output of a residual block. 
This mapping corresponds to a one-step Euler method for numerical integration.

Neural ODE extends this idea allowing us to use any numerical integration scheme:
\begin{equation}\label{eq:odenet}
        dz = f_\theta(z)dt \;\;\;\;\;\;\;\;\;\;\;\;\;\; z_{out} = ODESolver(z_{in}, f_\theta, t_{begin}, t_{end}),
\end{equation}
there $t_{begin}$ and $t_{end}$ are bounds of integration, $ODESolver$ is some numerical integration method.  Further, introduction a stochastic term to the differential equation leads to a neural SDE:
\begin{equation}\label{eq:sdenet}
        dz = f_\theta(z)dt + \sigma dW \;\;\;\;\;\;\;\;\; z_{out} = SDESolver(z_{in}, f_\theta, \sigma, t_{begin}, t_{end}),
\end{equation}
where $dW$ is the vector stochastic Wienner process of the same dimensionality as $z$, $\sigma$ is a scalar magnitude of stochasticity, $SDESolver$ is a numerical method for SDE integration. 

We designed models with a similar architecture to provide an objective and reliable comparison of residual networks, neural ODE, and neural SDE. Our neural networks basically consist of several sequences of blocks, which perform integration according to eq. (\ref{eq:resnet}), (\ref{eq:odenet}) or (\ref{eq:sdenet}).
These integration blocks are separated by down-sampling blocks. 
Therefore the main considered models are ResNet (with sequences of residual blocks), ODENet (with sequences of neural ODEs), SDENet (with sequences of neural SDEs). All models have the same architecture except for the type of integration blocks. Moreover, ResNet, ODENet and SDENet have similar functional form of a dynamic function $f_\theta$ inside the integration blocks.

\section{Experiments}
The main purpose of our experiments is to introduce stochasticity into continuous models and to investigate its regularisation properties. Additionally, we compare continuous models with a baseline residual network. As the first step in our study, we compare considered models on a toy task and observe that stochasticity performs as a good regularizer. Inspired by promising results, we continue comparison on CIFAR-10, CIFAR-100, and TinyImageNet.

Besides possible regularization properties, neural SDE provides an opportunity to improve quality with averaging predictions. Indeed, one can run trained SDENet $n$ times on test data and average predictions, obtained from different random trajectories.
Furthermore, it is possible to train a continuous model in a stochastic mode integrating eq.(\ref{eq:sdenet}) and switch to a deterministic mode during a test-time evaluation by replacing the integration scheme to eq.(\ref{eq:odenet}), which considers only deterministic dynamics $f_\theta$.
We explore these opportunities in our experiments.

It should be noted that neural SDE models contain $\sigma$ as a hyperparameter, which we choose using grid search (details of the search and chosen values can be found in Appendix).
Moreover, we use common back-propagation instead of adjoint method \cite{chen2018neural} to compute gradients during the training of continuous models, because of numerical instability of the adjoint method \cite{gholami2019anode}.

We published our code at\texttt{ https://github.com/AlexandraVolokhova/stochasticity\_in\_neural\_ode}

\subsection{Toy dataset}
 
We consider a binary classification task with samples from two 10-dimensional Gaussians. The distance between their centers equals 3.0, each Gaussian has an identity covariance matrix. This toy task has an optimal solution, which achieves 93.3\% accuracy.

The results are presented in Table \ref{tab:results_toy}. 
Our experiments show that ResNet and ODENet reach almost the same accuracy. In contrast to them, SDENet is able to achieve much better results reaching an almost optimal solution. It is interesting to note, that in the deterministic test-time mode, SDENet performs as well as averaging along 5-10 trajectories, but requiring less time to compute predictions.
\begin{table}[h]\small
  \caption{Test accuracy of considered models on the toy dataset. We repeat each experiment 5 times with different random seed and report mean $\pm$ standard deviation in percentages. SDENet\_0 denotes the deterministic test-time mode, SDENet\_n ($n>0$) denotes averaging predictions along $n$ stochastic trajectories during the test-time evaluation. 'Optimum' means the accuracy of the theoretical optimal solution}
  \centering
  \begin{tabular}{|c|c|c|c|c|c|c|c|c|}
    \toprule
    \cmidrule(r){1-2}
    ResNet&ODENet&SDENet\_0&SDENet\_1&SDENet\_2&SDENet\_5&SDENet\_10&SDENet\_20 & \textit{Optimum}\\
    \midrule
    $90.6 \pm 0.2$ & $90.8 \pm 0.9$ & $92.6 \pm 0.3$ & $91.2 \pm 0.2$ & $92.0 \pm 0.2$ & $92.5 \pm 0.3$ & $92.7 \pm 0.3$ & $92.8\pm 0.3$ & $93.3$\\
    \bottomrule
  \end{tabular}
  \label{tab:results_toy}
\end{table}
\subsection{CIFAR-10, CIFAR-100, and Tiny ImageNet}

\begin{wrapfigure}[18]{r}[0pt]{0.3\textwidth}
    \centering
    \includegraphics[width=\linewidth]{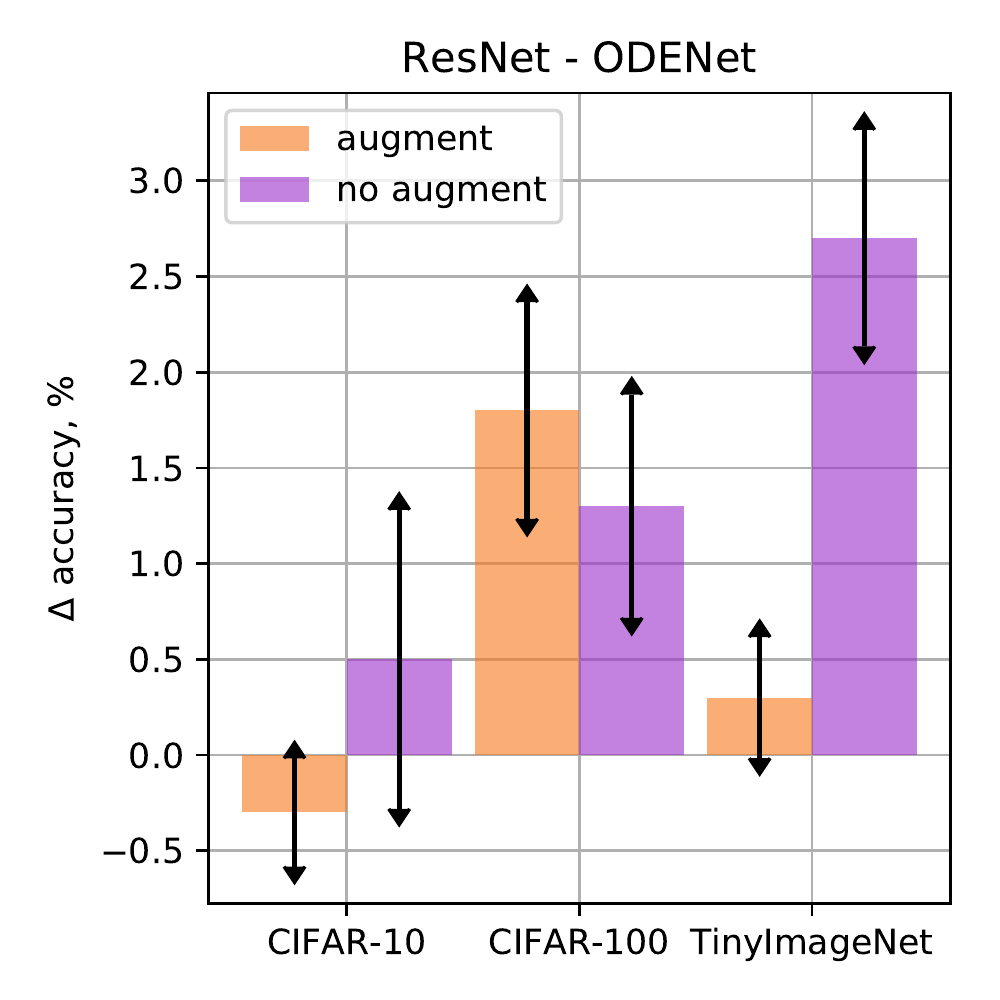}
    \caption{Difference between accuracy of ResNet and ODENet on three classification tasks. Colored bars show mean values of differences averaged by 5 runs, error bars demonstrate the standard deviation.}
    \label{fig:resnet_vs_odenet}
\end{wrapfigure}
\begin{wrapfigure}[17]{r}[0pt]{0.3\textwidth} 
    \centering
    \includegraphics[width=\linewidth]{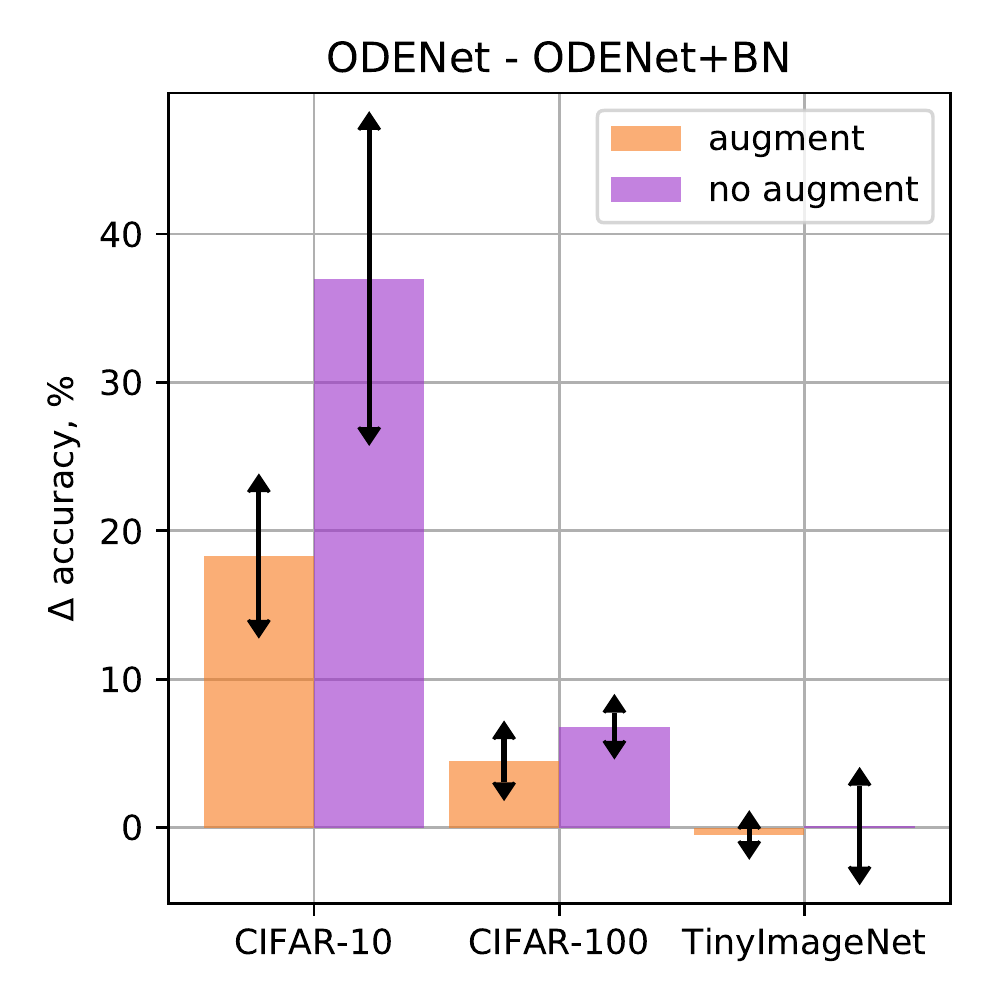}
    \caption{Difference between accuracy of ODENet and ODENet+BN on three classification tasks. Colored bars show mean values of differences averaged by 5 runs, error bars demonstrate the standard deviation.}
    \label{fig:odenet_bn_results}
\end{wrapfigure}

The considered models are trained on three image classification tasks: CIFAR-10, CIFAR-100, and TinyImageNet. 
We design ResNet, ODENet and SDENet similarly, except for the type of integration blocks, which are residual blocks, neural ODEs or neural SDEs (see Appendix, Fig.\ref{fig_3:architectures}). Integration is carried out inside ODENet and SDENet by the Runge-Kutta fourth-order method. We conduct our experiments with and without data augmentation in order to compare models in various conditions. The results of our experiments are presented in Table \ref{tab:results_appendix} in Appendix and on Figures \ref{fig:resnet_vs_odenet}, \ref{fig:sdenet_results}, \ref{fig:odenet_bn_results}.


\paragraph{ResNet v.s. ODENet}
According to our experiments, both model performs almost similarly,  with a occasional superiority of ResNet (see Fig.\ref{fig:resnet_vs_odenet} and Table \ref{tab:results_appendix} in Appendix).
Since the main difference between ResNet and ODENet is the numerical integration method (Euler v.s. fourth-order Runge-Kutta), we observe that more precise one does not improve the quality of classification. 
This result is quite reasonable because precise integration in the neural ODE restricts possible mappings to a set of homogeneous ones \cite{dupont2019augmented}, while rude integration in ResNet allows for breaking homogeneity. Therefore, taking into account \cite{dupont2019augmented} and our experiments we conclude that precise integration does not increase the expressiveness of the model.
\newpage
\paragraph{Regularization properties of a neural SDE}
We observe that the introduction of stochasticity into a neural ODE improves its generalization. 
SDENet consistently achieves better accuracy than ODENet in our experiments without augmentation (see Fig.\ref{fig:sdenet_results_noaug} and Table \ref{tab:results_appendix}). However, analogous experiments with augmentation show that neural ODE and neural SDE perform similarly (see Fig.\ref{fig:sdenet_results_aug} and Table \ref{tab:results_appendix}). Additionaly, ODENet with augmentation considerably outperforms SDENet without augmentation (see Table \ref{tab:results_appendix} and Fig.\ref{fig_2:results} in Appendix). Therefore, we conclude that stochasticity is actually able to regularize neural ODE, but simple data augmentation does it significantly better.
Moreover, an additional regularization effect from stochasticity is immaterial if data augmentation is used in the training procedure.

In addition, we observe that averaging predictions along stochastic trajectories improves performance. As one can see from Fig.\ref{fig:sdenet_results}, the quality of predictions continuously rises with an increase in the number of random trajectories in averaging. This result is reasonable because stochasticity is independently introduced to each trajectory, therefore averaging along them is able to reduce variance term of expected generalization error.
Moreover, switching to the deterministic mode at test-time performs mostly as averaging along several trajectories (this mode referred to as SDENet\_0 at Fig.\ref{fig:sdenet_results}). However, SDENet\_0 significantly loses quality in experiments on TinyImageNet with augmentation. We assume this effect can be explained as follows. The prediction averaged along the trajectories is an unbiased estimation of the prediction expected over the trajectories. Also, the deterministic trajectory in neural ODE is the expected trajectory in corresponding neural SDE due to the properties of the Wienner process. As far as the mapping from a trajectory to a prediction is a non-linear function, then the prediction based on the deterministic trajectory is a biased estimation to the expected prediction. The value of this bias depends on the variance of trajectories and properties of the non-linear function, that sometimes may lead to drops in the quality at the test time. For this reason, we recommend using the deterministic test-time mode with caution for neural SDE.
\begin{figure}
\begin{subfigure}{.5\textwidth}
  \centering
  \includegraphics[width=\linewidth]{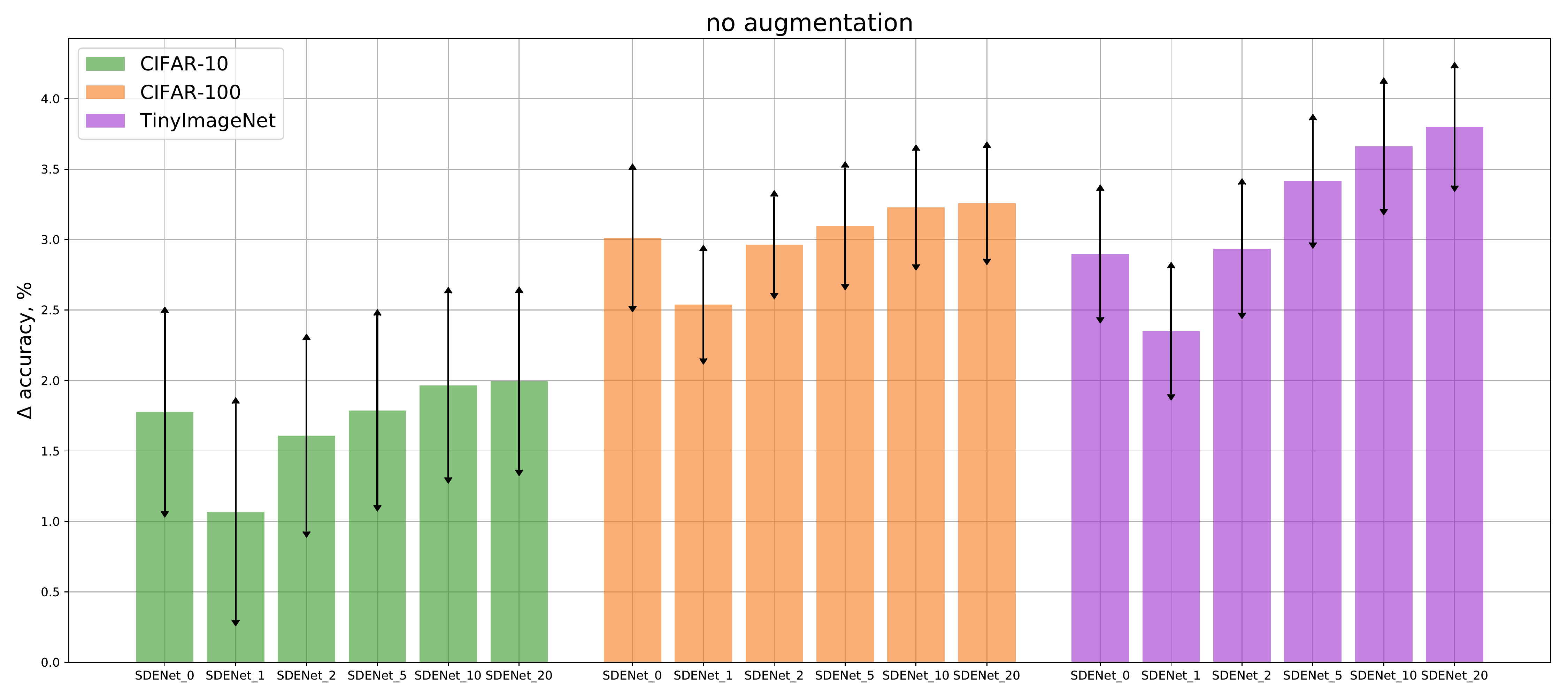}
  \caption{}
  \label{fig:sdenet_results_noaug}
\end{subfigure}
\begin{subfigure}{.5\textwidth}
  \centering
  \includegraphics[width=\linewidth]{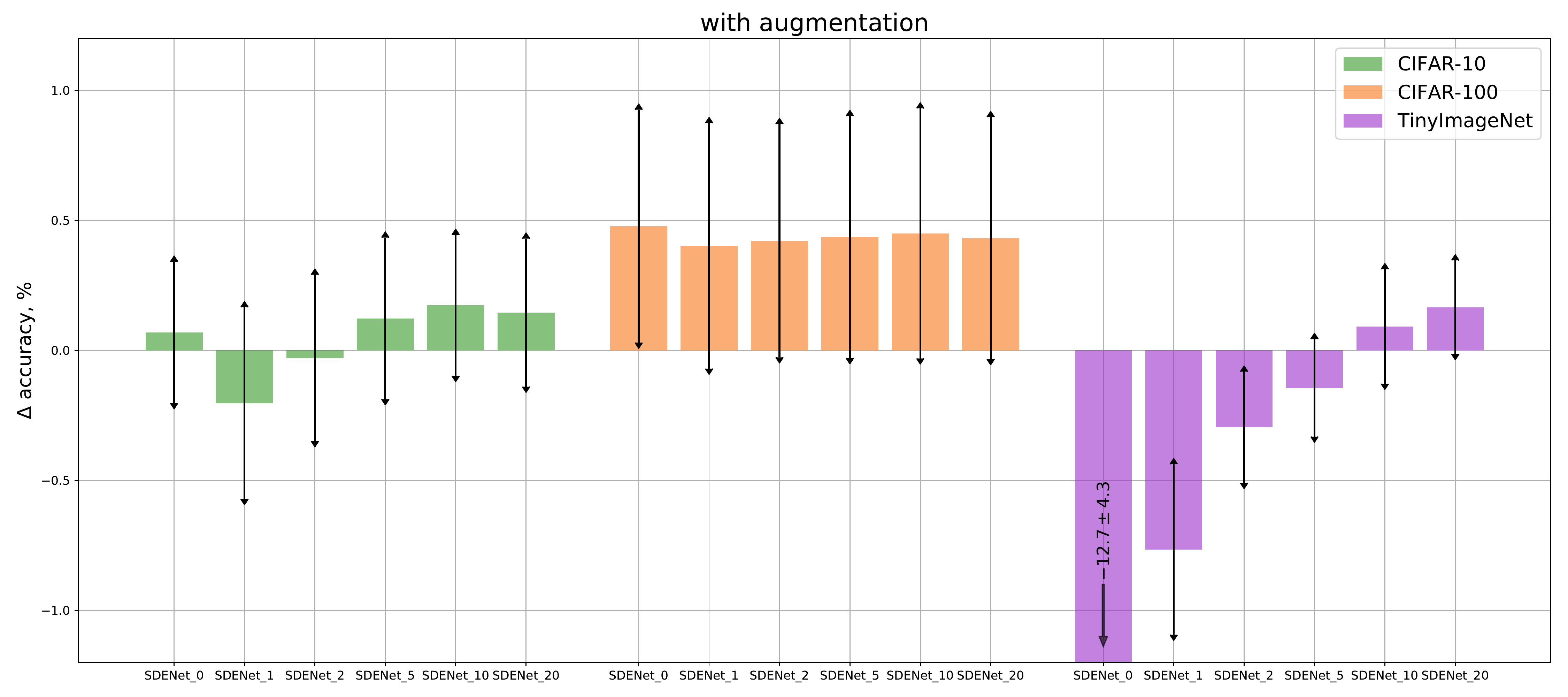}
  \caption{}
  \label{fig:sdenet_results_aug}
\end{subfigure}
\caption{Difference between accuracy of SDENet and ODENet on three classification tasks. Figures \ref{fig:sdenet_results_noaug} and \ref{fig:sdenet_results_aug} present results of experiments in settings without augmentation and with augmentation correspondingly.  Colored bars show mean values of differences averaged by 5 runs, error bars demonstrate the standard deviation. SDENet\_0 denotes the deterministic test-time mode, SDENet\_n ($n>0$) denotes averaging predictions along $n$ stochastic trajectories during the test-time evaluation. There is a significant negative value for the difference between SDENet\_0 and ODENet on TinyImageNet with augmentation. We don't fully depict the corresponding bar for illustrative reasons.}
\label{fig:sdenet_results}
\end{figure}

\paragraph{Batch normalization inside a neural ODE}\label{bn_text}
Batch normalization \cite{ioffe2015batch} is a very common technique in modern neural networks. 
However, there is a significant difference between applying batch normalization in residual networks and neural ODEs. 
Common implementations of residual networks usually contain batch normalization inside residual blocks.
However, if we put batch normalization into the dynamic function $f_\theta$ of neural ODE, then the same normalization will be applied to internal representations $z(t)$ at different time points $t$ during numerical integration. In this case, moving averages of batch normalization will be accumulated along all steps of numerical integration. It is not clear how this affects the performance of neural ODEs. 
Therefore, we design ODENet and SDENet without batch normalization inside their dynamic functions. Additionally, we train ODENet with batch normalization inside its dynamic functions, which we refer to as ODENet+BN.
Our experiments show that neural ODE with batch normalization occasionally works significantly worse than without batch normalization, which confirms our apprehension (see Fig.\ref{fig:odenet_bn_results} and Table \ref{tab:results_appendix}). However, we suppose this effect should be investigated more thoroughly in the future.

\section{Conclusion}
We present an empirical study of neural ODEs and neural SDEs on various classification tasks. 
The main contribution of this paper is the exploration of regularization properties of neural SDE. We find out that stochastic term in the neural differential equation allows us to increase generalization of the model if we train it without data augmentation. However, when the model learns in a setting with data augmentation, additional stochasticity of the differential equation does not increase the quality. Hence, we conclude that stochasticity is not enough powerful regularizer for neural ODE in case of image classification. Nonetheless, neural SDE may be able to significantly improve quality on tasks, where data augmentation is hard to handle.

In addition, we compare the performance of continuous models and residual networks. We observe that ResNet occasionally works better than neural ODE on image classification tasks. That means more accurate integration, which is performed in the neural ODE, does not increase expressiveness. It is worth to note that more precise integration requires more computations, which leads to longer training and inference procedure. Taking into account our experiments, we would not recommend using continuous models for image classification tasks, as a residual network manages them better and more efficiently. Continuous models seem to be more applicable to time-series generative models, as it is reported in \cite{rubanova2019latentode, li2020scalablesde}.

\section*{Acknowledgments}
This research is in part based on the work supported by Samsung Research, Samsung Electronics. Results on stochasticity in neural ODE have been supported by the Russian Science Foundation grant no.17-71-20072. This research was supported in part through computational resources of HPC facilities at NRU HSE. We sincerely thank Kirill Neklyudov and Arsenii Ashukha for their sound advice. Special thanks to Kirill for his considerable assistance in writing the text of this paper. 
\bibliographystyle{unsrt}  
\bibliography{references}  

\begin{thebibliography}{10}

\bibitem{murphy2012machine}
Kevin~P Murphy.
\newblock {\em Machine learning: a probabilistic perspective}.
\newblock MIT press, 2012.

\bibitem{srivastava2014dropout}
Nitish Srivastava, Geoffrey Hinton, Alex Krizhevsky, Ilya Sutskever, and Ruslan
  Salakhutdinov.
\newblock Dropout: a simple way to prevent neural networks from overfitting.
\newblock {\em The journal of machine learning research}, 15(1):1929--1958,
  2014.

\bibitem{ioffe2015batch}
Sergey Ioffe and Christian Szegedy.
\newblock Batch normalization: Accelerating deep network training by reducing
  internal covariate shift.
\newblock {\em arXiv preprint arXiv:1502.03167}, 2015.

\bibitem{chen2018neural}
Tian~Qi Chen, Yulia Rubanova, Jesse Bettencourt, and David~K Duvenaud.
\newblock Neural ordinary differential equations.
\newblock In {\em Advances in neural information processing systems}, pages
  6571--6583, 2018.

\bibitem{grathwohl2018ffjord}
Will Grathwohl, Ricky~TQ Chen, Jesse Bettencourt, Ilya Sutskever, and David
  Duvenaud.
\newblock Ffjord: Free-form continuous dynamics for scalable reversible
  generative models.
\newblock {\em arXiv preprint arXiv:1810.01367}, 2018.

\bibitem{rubanova2019latentode}
Yulia Rubanova, Tian~Qi Chen, and David~K Duvenaud.
\newblock Latent ordinary differential equations for irregularly-sampled time
  series.
\newblock In {\em Advances in Neural Information Processing Systems}, pages
  5321--5331, 2019.

\bibitem{yildiz2019ode2vae}
Cagatay Yildiz, Markus Heinonen, and Harri Lahdesmaki.
\newblock Ode2vae: Deep generative second order odes with bayesian neural
  networks.
\newblock In {\em Advances in Neural Information Processing Systems}, pages
  13412--13421, 2019.

\bibitem{jia2019neuraljumpsde}
Junteng Jia and Austin~R Benson.
\newblock Neural jump stochastic differential equations.
\newblock In {\em Advances in Neural Information Processing Systems}, pages
  9843--9854, 2019.

\bibitem{li2020scalablesde}
Xuechen Li, Ting-Kam~Leonard Wong, Ricky~TQ Chen, and David Duvenaud.
\newblock Scalable gradients for stochastic differential equations.
\newblock {\em arXiv preprint arXiv:2001.01328}, 2020.

\bibitem{he2016deep}
Kaiming He, Xiangyu Zhang, Shaoqing Ren, and Jian Sun.
\newblock Deep residual learning for image recognition.
\newblock In {\em Proceedings of the IEEE conference on computer vision and
  pattern recognition}, pages 770--778, 2016.

\bibitem{tzen2019theoreticalsde}
Belinda Tzen and Maxim Raginsky.
\newblock Theoretical guarantees for sampling and inference in generative
  models with latent diffusions.
\newblock {\em arXiv preprint arXiv:1903.01608}, 2019.

\bibitem{liu2019neuralsde}
Xuanqing Liu, Tesi Xiao, Si~Si, Qin Cao, Sanjiv Kumar, and Cho-Jui Hsieh.
\newblock Neural sde: Stabilizing neural ode networks with stochastic noise.
\newblock {\em arXiv preprint arXiv:1906.02355}, 2019.

\bibitem{tzen2019neuralsde}
Belinda Tzen and Maxim Raginsky.
\newblock Neural stochastic differential equations: Deep latent gaussian models
  in the diffusion limit.
\newblock {\em arXiv preprint arXiv:1905.09883}, 2019.

\bibitem{krizhevsky2009cifar}
Alex Krizhevsky, Geoffrey Hinton, et~al.
\newblock Learning multiple layers of features from tiny images.
\newblock 2009.

\bibitem{tinyimagenet}
{\em \url{https://tiny-imagenet.herokuapp.com}}.

\bibitem{gholami2019anode}
Amir Gholami, Kurt Keutzer, and George Biros.
\newblock Anode: Unconditionally accurate memory-efficient gradients for neural
  odes.
\newblock {\em arXiv preprint arXiv:1902.10298}, 2019.

\bibitem{dupont2019augmented}
Emilien Dupont, Arnaud Doucet, and Yee~Whye Teh.
\newblock Augmented neural odes.
\newblock In {\em Advances in Neural Information Processing Systems}, pages
  3134--3144, 2019.

\bibitem{paszke2017automatic}
Adam Paszke, Sam Gross, Soumith Chintala, Gregory Chanan, Edward Yang, Zachary
  DeVito, Zeming Lin, Alban Desmaison, Luca Antiga, and Adam Lerer.
\newblock Automatic differentiation in pytorch.
\newblock 2017.

\end{thebibliography}





\newpage
\section*{Appendix}\label{sec:appendix}
This section provides all the details of our experiments with CIFAR-10, CIFAR-100, and TinyImageNet, which are necessary for the possible reproduction of our results.
\subsection*{Detailed results of the main experiments}
\begin{figure}[h]
\includegraphics[width=1\linewidth]{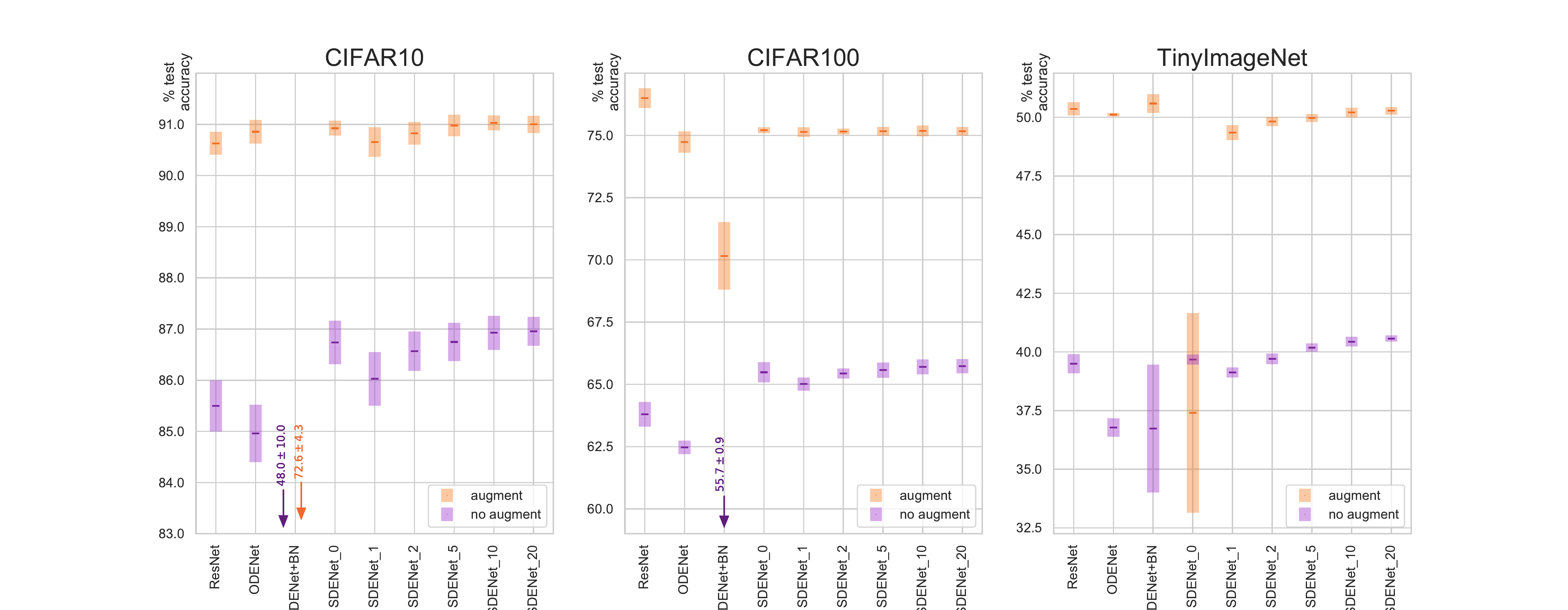}
\caption{Illustrative representation of the figures from Table \ref{tab:results_appendix}. Mean values are depicted as bold line segments and standard deviations are depicted as half-transparent rectangles. Outlying  values are not shown for visualization reasons.}
\label{fig_2:results}
\end{figure}

Tabel \ref{tab:results_appendix} contains test accuracies of considered models on image classification tasks. Fig.\ref{fig_2:results} illustrates these results for easier comparison.
\begin{table}[h]
 \caption{Test accuracy of considered models on three classification tasks in different conditions. Columns 'augment' and 'no augment' contain results of experiments with augmentation and without it respectively. We repeat each experiment 5 times with different random seeds and report the mean $\pm$ standard deviation. SDENet\_0 denotes the deterministic test-time mode, SDENet\_n ($n>0$) denotes averaging predictions along $n$ stochastic trajectories during the test-time evaluation.
 }
  \centering
  \begin{tabular}{|l|c|c|c|c|c|c|}
    \toprule
    & \multicolumn{2}{| c |}{CIFAR-10} & \multicolumn{2}{| c |}{CIFAR-100} & \multicolumn{2}{| c |}{TinyImageNet} \\
    \cmidrule(r){2-7}
    & augment & no augment & augment & no augment & augment & no augment \\
     \midrule
     ResNet & $90.6 \pm 0.2$ & $85.5 \pm 0.5$ & $76.5 \pm 0.4$ & $63.8 \pm 0.5$ & $50.4 \pm 0.3$ & $39.5 \pm 0.4$\\
     ODENet & $90.9 \pm 0.2$ & $85.0 \pm 0.6$ & $74.7 \pm 0.4$ & $62.5 \pm 0.3$ & $50.1 \pm 0.1$ & $36.8 \pm 0.4$\\
     ODENet+BN & $72.6 \pm 4.3$ & $48.0 \pm 10.0$ & $70.2 \pm 1.4$ & $55.7 \pm 0.9$ & $50.6 \pm 0.4$ & $36.7 \pm 2.7$\\
     SDENet\_0 & $90.9 \pm 0.1$ & $86.7 \pm 0.4$ & $75.2 \pm 0.1$ & $65.5 \pm 0.4$ & $37.4 \pm 4.3$ & $39.7 \pm 0.2$\\
     SDENet\_1 & $90.7 \pm 0.3$ & $86.0 \pm 0.5$ & $75.1 \pm 0.2$ & $65.0 \pm 0.3$ & $49.4 \pm 0.3$ & $39.1 \pm 0.2$\\
     SDENet\_2 & $90.8 \pm 0.2$ & $86.6 \pm 0.4$ & $75.2 \pm 0.1$ & $65.4 \pm 0.2$ & $49.8 \pm 0.2$ & $39.7 \pm 0.2$\\
     SDENet\_5 & $91.0 \pm 0.2$ & $86.7 \pm 0.4$ & $75.2 \pm 0.2$ & $65.6 \pm 0.3$ & $50.0 \pm 0.2$ & $40.2 \pm 0.2$\\
     SDENet\_10 & $91.0 \pm 0.1$ & $86.9 \pm 0.3$ & $75.2 \pm 0.2$ & $65.7 \pm 0.3$ & $50.2 \pm 0.2$ & $40.4 \pm 0.2$\\
     SDENet\_20 & $91.0 \pm 0.2$ & $87.0 \pm 0.3$ & $75.2 \pm 0.2$ & $65.7 \pm 0.3$ & $50.3 \pm 0.2$ & $40.6 \pm 0.1$\\
    \bottomrule
  \end{tabular}
  \label{tab:results_appendix}
\end{table}

\subsection*{Hyperparameters and training procedure}
We divide the training dataset into train and validation parts for hyperparameter search. After that, we choose hyperparameters, with those the model achieves better accuracy on the validation part.  Finally, we join divided parts and train the final model on the full training dataset. The main hyperparameters, which we selected in this way, are $\sigma$, learning rate, and batch size.

We conduct our experiments using PyTorch \cite{paszke2017automatic} and torchdiffeq library \cite{chen2018neural}. 
We use stochastic gradient descent with Nesterov momentum equals 0.9 and learning rate scheduler \texttt{ReduceLROnPlateau} with \texttt{threshold\_mode='rel'}, \texttt{threshold=0.02}, \texttt{patience=10}, \texttt{factor=0.1}, \texttt{min\_lr=1e-5}. 
Also, we  use \texttt{WarmUpLR} with different \texttt{warmup\_steps}. Learning rate (lr), warmup\_steps (warm) and weight decay (wd) are different for different experiments (see Table \ref{tab:param_noaug} and \ref{tab:param_aug}). Other heperparameters are shown in the Table \ref{tab:param_datasets}.

\begin{table}[h]
 \caption{Optimizer parameters no augmentation}
  \centering
  \begin{tabular}{|l|c|c|c|c|c|c|c|c|c|}
    \toprule
    & \multicolumn{3}{| c |}{CIFAR-10} & \multicolumn{3}{| c |}{CIFAR-100} & \multicolumn{3}{| c |}{TinyImageNet} \\
    \cmidrule(r){2-10}
    & lr & warm & wd & lr & warm & wd & lr & warm & wd\\
     \midrule
     ResNet & 0.4 & 0 & 5e-4 & 0.4 & 2 & 5e-4 & 0.05 & 3 & 1e-4\\
     ODENet & 0.1 & 0 & 5e-4 & 0.1 & 2 & 5e-4 & 0.05 & 0 & 1e-5\\
     ODENet+BN & 0.05 & 0 & 5e-4 & 0.1 & 2 & 5e-4 & 0.1 & 0 & 1e-5\\
     SDENet & 0.05 & 0 & 5e-4 & 0.1 & 2 & 5e-4 & 0.05 & 0 & 1e-5\\
    \bottomrule
  \end{tabular}
  \label{tab:param_noaug}
\end{table}
\FloatBarrier
\begin{table}[h]
 \caption{Optimizer parameters with augmentation}
  \centering
  \begin{tabular}{|l|c|c|c|c|c|c|c|c|c|}
    \toprule
    & \multicolumn{3}{| c |}{CIFAR-10} & \multicolumn{3}{| c |}{CIFAR-100} & \multicolumn{3}{| c |}{TinyImageNet} \\
    \cmidrule(r){2-10}
    & lr & warm & wd & lr & warm & wd & lr & warm & wd\\
     \midrule
     ResNet & 0.1 & 0 & 5e-4 & 0.2 & 2 & 5e-4 & 0.1 & 3 & 1e-4\\
     ODENet & 0.05 & 0 & 5e-4 & 0.1 & 2 & 5e-4 & 0.01 & 3 & 1e-4\\
     ODENet+BN & 0.05 & 0 & 5e-4 & 0.1 & 2 & 5e-4 & 0.05 & 3 & 1e-4\\
     SDENet & 0.05 & 0 & 5e-4 & 0.1 & 2 & 5e-4 & 0.01 & 3 & 1e-4\\
    \bottomrule
  \end{tabular}
  \label{tab:param_aug}
\end{table}
\FloatBarrier
\begin{table}[h]
 \caption{Other hyperparameters. \# steps denotes number of steps of numerical integration}
  \centering
  \begin{tabular}{|l|c|c|c|}
  \toprule
    \cmidrule(r){1-2}
    & CIFAR-10 & CIFAR-100 & TinyImageNet\\
     \midrule
     batch size & 512 & 256 & 256\\
     \# steps for ODENet and SDENet & 10 & 3 & 2\\
     \# steps for ODENet+BN & 6 & 3 & 2\\
     $\sigma$ (no augment) & 0.79 & 0.25 & 0.4\\
     $\sigma$ (with augment) & 0.2 & 0.05 & 0.3\\
    \bottomrule
  \end{tabular}
  \label{tab:param_datasets}
\end{table}

It is important to note that TinyImageNet from \cite{tinyimagenet} initially consists of three sets: train, validation, and test. Train and validation sets are labeled and the test set does not have labels. Therefore, we use only the train set to train our models and validation set for the final evaluation.

\subsection*{Architectures}

\begin{figure}
\includegraphics[width=1\linewidth]{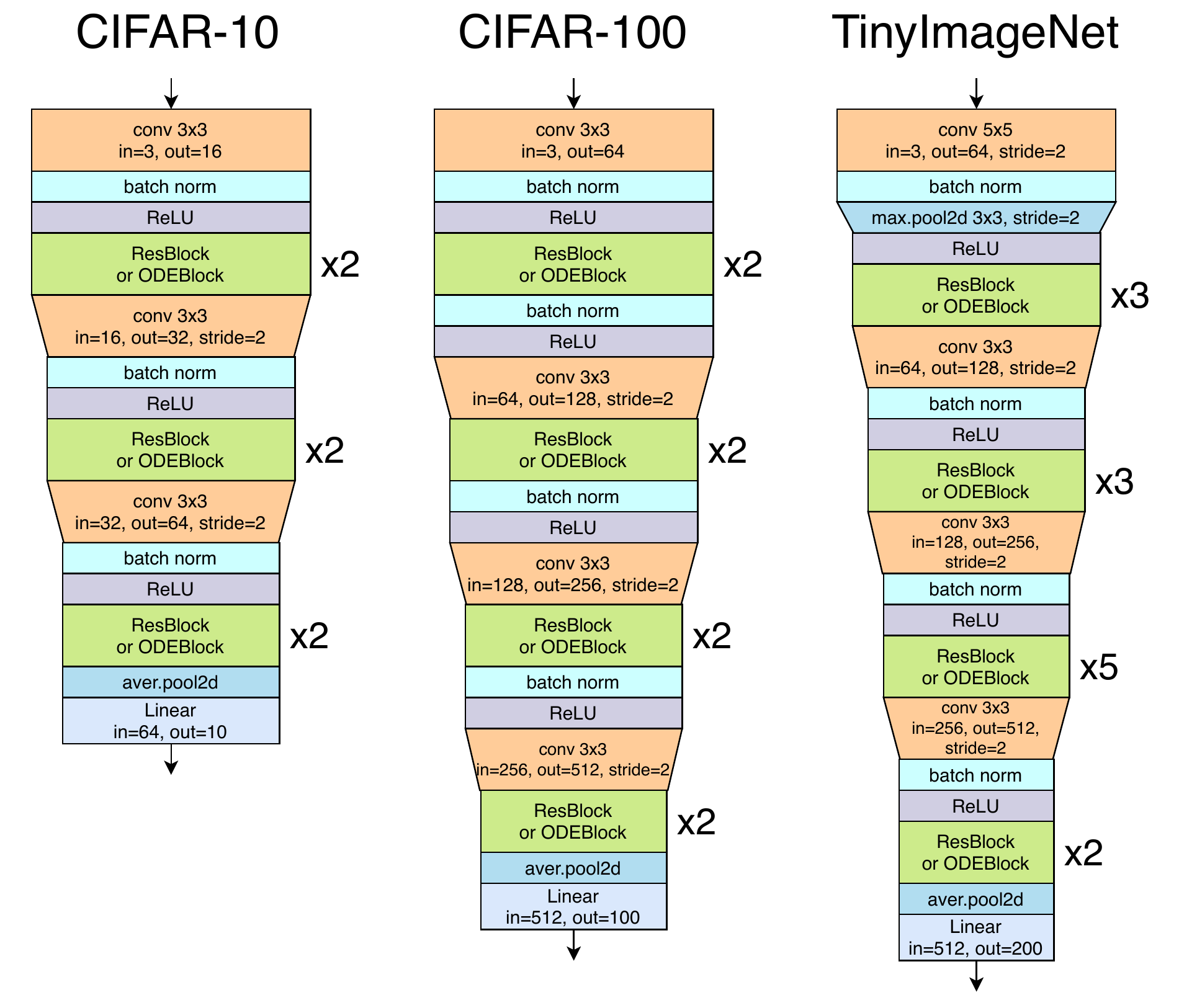}
\caption{Architectures of models. $\times \;n$ near a block means that it is repeated $n$ times}
\label{fig_3:architectures}
\end{figure}

\begin{figure}
\centering
\includegraphics[width=0.8\linewidth]{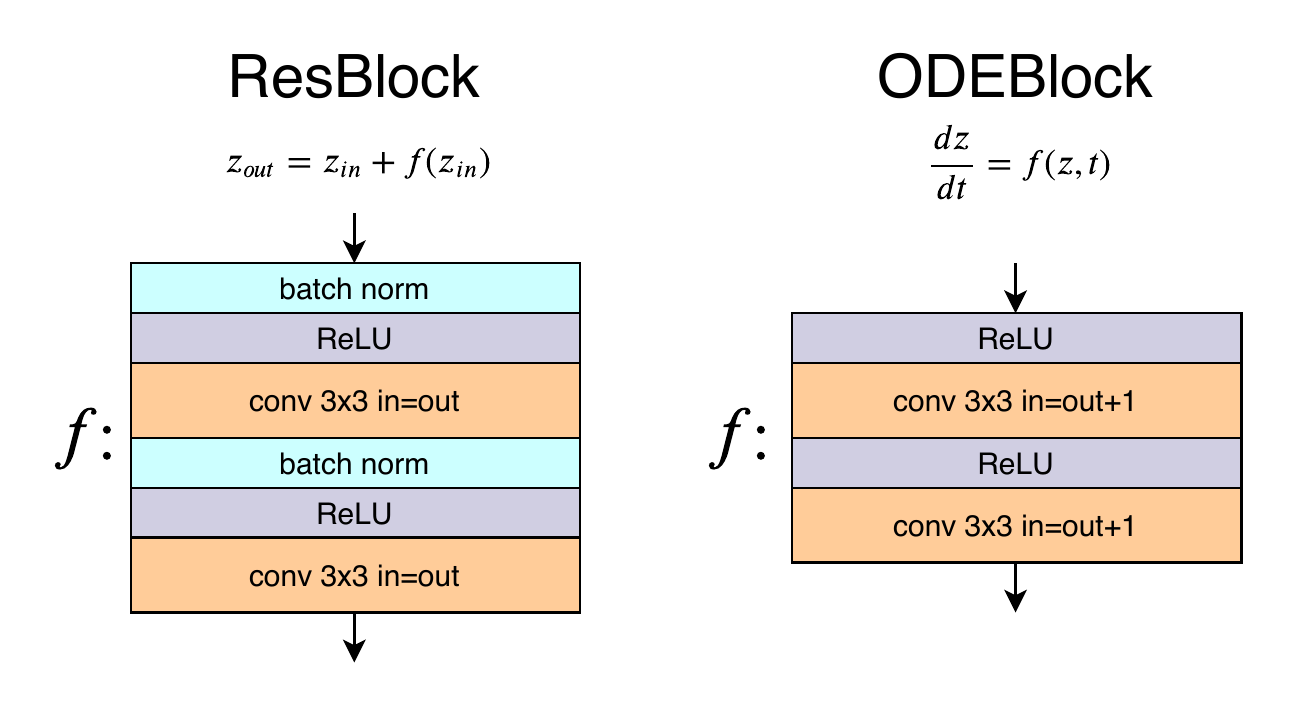}
\caption{Integration blocks}
\label{fig_4:blocks}
\end{figure}

Figures \ref{fig_3:architectures} and \ref{fig_4:blocks} depict architectures of our models for each dataset. We design our models to be strong enough to achieve adequate accuracy on the test set and 100\% accuracy on the training set. Hence, our models are able to overfit it is reasonable to study regularizers with them. 

We design residual networks and continuous models in a similar way, so ResNet differs from ODENet and SDENet only in  the type of integration block see Fig.\ref{fig_4:blocks}). ResBlock is used in ReNet, ODEBlock is used in ODENet and SDENet. 

Continuous models have the same architecture, but SDENet differs from ODENet in the stochastic term in eq.\ref{eq:sdenet}. So, these models differ also in solvers of neural differential equations.

As we mentioned in the section \ref{bn_text}, we do not put batch normalization into dynamic function $f$ of ODENet and SDENet. However, in our experiment with ODENet+BN, we add batch normalization layers into $f$ in the same way as we do for ResBlock.

Additionally, dynamic function $f$ in ODEBlock depends on time $t$. We design this dependency adding one extra channel to the internal representation $z(t)$. This extra channel is filled by the value of time $t$. Therefore, convolution layers have \texttt{input\_channels = output\_channels + 1} in ODEBlock.
\FloatBarrier





\end{document}